\documentclass[11pt]{article}
\usepackage[T1]{fontenc}
\usepackage{amsthm, amsmath, amssymb, amsfonts, url, booktabs, tikz, setspace, fancyhdr, bm, mathrsfs}
\usepackage{geometry}
\usepackage{hyperref, enumerate}
\usepackage[shortlabels]{enumitem}
\usepackage[babel]{microtype}
\usepackage[english]{babel}
\usepackage[capitalise]{cleveref}
\usepackage{comment}
\usepackage{bbm}
\usepackage{csquotes}
\usepackage{cite}
\usepackage{mathabx}
\usepackage{graphicx}
\usepackage{float}
\usepackage{subcaption}

\counterwithin{figure}{section}

\theoremstyle{definition}

\newtheorem*{defn-non}{Definition}

\newlist{Case}{enumerate}{3}
\setlist[Case, 1]{%
    label           =   {\bfseries Case \arabic*.},
    labelindent=1em ,labelwidth=1cm, labelsep*=1em, leftmargin =!
}
\setlist[Case, 2]{%
    label           =   {\bfseries Subcase \arabic{Casei}.\arabic*.},
    labelindent=-1em ,labelwidth=1cm, labelsep*=1em, leftmargin =!
}
\setlist[Case, 3]{%
    label           =   {\bfseries Subsubcase \arabic{Casei}.\arabic{Caseii}.\arabic*.},
    labelindent=-1em ,labelwidth=1cm, labelsep*=1em, leftmargin =!
}

\usepackage{todonotes}

\title{Understand the Effect of Importance Weighting in Deep Learning on Dataset Shift}

\author{
Vo Thien Nhan \thanks{
Institute of Engineering, Ho Chi Minh City University of Technology (HUTECH), Vietnam \\  Email: thiennhan.math@gmail.com}}

\begin{document}
\maketitle

\begin{abstract}
Deep neural networks have demonstrated remarkable flexibility and performance across a variety of prediction tasks, yet the efficacy of classical importance weighting techniques under dataset shift remains insufficiently understood in this context. In this project, we systematically investigate the interaction between importance‑weighted loss functions and deep network training dynamics under both label shift and covariate shift scenarios. Following the framework of Byrd \& Lipton \cite{ref1}, we first conduct experiments on synthetic two‑dimensional datasets—linearly separable and moon‑shaped distributions—using logistic regression and multilayer perceptrons to visualize decision boundary evolution and assess the persistence of weighting effects. We then extend our analysis to image classification tasks on CIFAR‑10, evaluating binary classifiers under varying class imbalance ratios and training regimes with and without \(L_2\) regularization or dropout. Finally, we design a covariate shift experiment by re‑casting cat/dog and car/truck classes into higher‑level animal/vehicle categories to probe the limits of weighting strategies in deep convolutional networks. Our findings corroborate previous results: importance weighting significantly influences model behavior during early training but dissipates with prolonged optimization; \(L_2\) regularization, but not dropout, partially restores the weighting effect in deep nets; and, while weighting effectively corrects imbalance in simple models, it does not yield consistent benefits for over‑parameterized networks on CIFAR‑10. Covariate shift experiments reveal negligible performance gains, suggesting inherent challenges in applying instance‑level weighting when feature distributions diverge in complex data. These insights call into question the utility of importance weighting for deep learning practitioners facing real‑world distributional shifts.

\end{abstract}

\section{Introduction}
Importance weighting is an essential method in statistics and machine learning to estimate a quantity on some target distribution, but can only sample from a different source distribution. Though deep learning becomes to dominate a broad set of prediction tasks, the effect of importance weighting over deep neural networks is little known. In our project, we try to leverage the paper \cite{ref1} to investigate how importance weighting interact with deep neural networks.

We start by following the paper \cite{ref1} to conduct a series of experiments across a variety of architectures. We train Logistic Regression and MLP on synthetic 2-dimensional toy datasets and visualize the change of their decision boundaries in training process, in order to compare the impact of importance weighting on different models in different scenarios. We also train the self-defined convolutional neural networks (CNNs) for binary classification task on CIFAR-10 images. In this process, we investigate the effects of importance weighting on deep neural networks on label shift over the course of training. We train the models with and without L2 regularization and dropout. Aside from this, we also test the effects of importance weighting for class imbalance by sub-sampling CIFAR-10 training examples with different class ratios.

We compare our results with the paper~\cite{ref1}. The paper concludes that for standard neural networks, weighting has a significant effect early in training. However, as training progresses, the effect dissipates. We find the claims from the paper is consistent with the results of our experiments. Our experiment results also agree with the paper that L2-regularization but not dropout restores certain effects of importance weight, but it only holds true in the case of balanced training data. In general, consistent with the conclusions in the paper, our results call into question the application of importance weighting in the context of deep networks.

\section{Related Work} 
Importance weighted risk minimization is a standard tool in many machine learning tasks~\cite{ref2, ref3}. To formalize it rigorously, given $n$ samples from a source distribution $q(x)$, the task is to estimate some function of data $f(x)$ with samples from the target distribution $p(x)$. An unbiased estimator can be achieved using importance weighting
\begin{align}
\mathbb{E}_q\left[\frac{p(x)}{q(x)} f(x)\right] = \int_{x} f(x) \frac{p(x)}{q(x)} q(x) dx = \mathbb{E}_p[f(x)]
\end{align}

In the situation of domain adaptation, there are two types of data shift that often happen in real world dataset and can be adjusted by importance weighting. One is covariate shift where training samples $P_{\text{train}}(x)$ and testing samples $P_{\text{test}}(x)$ are from two distributions~\cite{ref4, ref5}. The other is label shift where the distribution of training and testing labels $P_{\text{train}}(y)$, $P_{\text{test}}(y)$ are different~\cite{ref6}.

Several works also try to incorporate importance weighting into deep neural networks. The work~\cite{ref7} uses weighted loss function to estimate individual treatment effects with different treated and control distributions. The importance-weighted risk minimization has been adopted in deep networks to combat label shifts~\cite{ref6, ref8}. Importance weighting has also been widely applied in deep reinforcement learning. It is used to learn from logged contextual bandit feedback~\cite{ref9} and applied as weighted sampling for performing TD updates~\cite{ref10}.
\section{Experimental Design}
\subsection{Label Shift}
\textbf{Synthetic Data} To validate the effect of importance weights during the training progress, we generate the 2-D toy data and visualize the results as described in the paper~\cite{ref1}. We first sample 512 points from truncated multinomial distribution with mean $[0, 0]$ and covariance matrix $\mathbf{I}_2$, and denote these points with positive labels. Then we rotate and translate the points to create the 512 negative samples, and make sure this data set is linearly separable. We generate test data in the same way.

We train a Logistic Regression and a Multi-Layer Perceptron on the synthetic dataset. The MLP has a single hidden layer with 64 units and use ReLU as activation function. Both models are trained by the SGD optimizer with a batch size of 8 for 100000 epochs, and the learning rate is $\frac{0.01}{\sigma_{\text{max}}(X)}$, where $\sigma_{\text{max}}(X)$ is the maximum singular value of the data matrix. The learning rate is around $1\text{e} - 04$ in our scenario. We train both models with and without L2 regularization in different experiments, and try an extra experiment with MLP using dropout without regularization.

We also experiment on the 2-D moon dataset, which is not linearly separable. We generate 1024 samples evenly distributed on positive and negative classes, and split it into training set and testing set. By applying the Logistic Regression and the Multi-Layer Perceptron, we explore how the decision boundary changes with inappropriate or appropriate models. Additionally, we generate imbalanced training set for the moon data by sub-sampling with positive to negative ratios $r \in \{10:1, 1:10\}$. For each imbalanced training set, we train LR and MLP with loss function weighted by $\frac{1}{r}$ and with unweighted loss function to show the impact of importance weights on imbalanced training examples.

\textbf{CIFAR-10 Binary Classification} We apply class-conditioned weights of various strengths on the binary classification of CIFAR-10 images to evaluate the impact of the weights on the learned decision boundaries. The CIFAR-10 are labeled subsets of the 80 million tiny images dataset. The dataset consists of 60000 color images of size $32 \times 32$ in 10 classes, with 6000 images per class. There are 50000 training images and 10000 test images. We train a binary classifier on training images labeled as cats or dogs (5000 per class). We use a simple convolutional neural network following the paper~\cite{ref1}: two convolution layers with 64 3×3 filters each and stride 1, followed by a 2×2 max pooling layer, followed by three convolution layers with 128 3×3 filters each and stride 1, followed by a second 2×2 max pooling layer, followed by two dense layers with 512 and 128 hidden units respectively and finally a dense layer to give binary outputs. All hidden layers employ ReLU activation functions. Two dropout layers are applied before each dense layer for models using dropout.

The models are trained for 500 epochs using minibatch SGD with a batch size of 16 and momentum as 0.9. The learning rate for SGD is 0.01. We evaluate on all 10000 test images from all classes, as well as 1000 random noise images. Experiments were run with importance weights $w \in \{1 : 128, 1 : 32, 1 : 8, 1 : 1, 8 : 1, 32 : 1, 128 : 1\}$. We also run experiments using the CNN classifier with the L2 regularization and the dropout. For L2 regularization, we set the penalty coefficient as 0.001. For the dropout models, we set the dropout rate as 0.5. In order to compare the agreement between models with different importance weights in different test distribution, we compute the fraction of images classified as dogs in the cat and dog classes, other 8 classes, and random noise images separately. Each model runs with 3 different random seeds to consider the standard deviation.

\textbf{CIFAR-10 Binary Imbalanced} We also conduct experiments to investigate the effect of importance weighting on class imbalance. We train a binary classifier for two classes of CIFAR-10: cat and dog. To simulate the class imbalance situation, we sub-sample the dog and cat training examples with ratios of $r \in \{16:1, 8:1, 4:1, 1:4, 1:8, 1:16\}$. For each ratio, we train models without importance weighting and with loss function weighted by $1/r$. We also train models with weighted loss function and \textit{L2} regularization. We use the same CNN architecture as defined above and hyperparameters also hold the same. For each experiment, we train 3 models with a different random seed to compute mean and standard deviation.
\subsection{Covariate Shift}
Having noticed the effects of the importance weight on the label shift problem, we became interested in testing its impacts on the covariate shift problem. Unlike the label shift, which we could easily create by making the training and testing datasets imbalanced, the covariate shift is hard to quantify in the image classification scenario. Theoretically, each pixel of each channel serves as a feature of an image, and we would need to specify the joint distribution of all the pixels in both training and testing datasets to measure the covariate shift and come up with sensible importance weights. However, given the infinite number of values each pixel could take, it is extremely challenging for us to even approximate such distribution. Therefore, we would like to produce the covariate shift on a higher level by combining the original 4 classes cat, dog, car and trunk into 2 new classes: animal and vehicle. Our new task became classifying the images of cat, dog, car or truck as one of the two new classes. Two images originally belonging to cat and dog respectively contain the separate information that resembles the information conveyed by two different pixels in the original image classification task. And the covariate shift originally bonded to the change of distribution of pixels was then realized as the different ratios of cat/dog (or car/trunk) in training and testing datasets. By introducing this abstraction, we were able to precisely define the importance weights to be the inverse of the cat/dog ratio (or similarly car/trunk ratio), and therefore study the impacts of the importance weight in the image classification scenario.

We anticipated the difficulties of our new image classification task. The crucial patterns for identifying a cat could be different from those of identifying a dog, making the neural networks struggle to discover common features. Therefore, besides the weighted model, we would also fit an unweighted model, and a model trained on a dataset with no covariate shift. We would expect our weighted model to outperform the unweighted model on the dataset with covariate shift but to underperform the model on the dataset with no covariate shift. As long as the importance weights make the performances closer to those resulted from the optimal setting (with no covariate shift), we would conclude the importance weights indeed played an role.
\section{Results}
\subsection{Synthetic Data}

\begin{figure}[H]
    \centering
    \includegraphics[width= 0.85\textwidth]{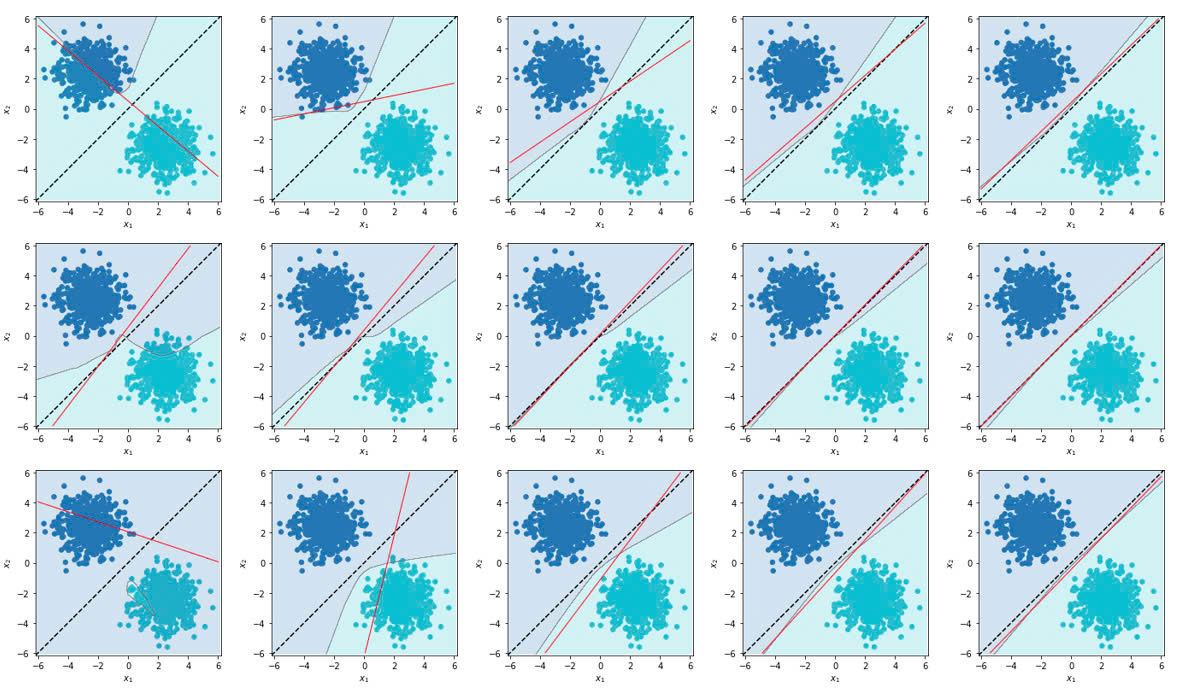}
    \caption{
        Results on linearly-separable dataset. The model is LR and MLP without L2 regularization. 
        From top to bottom, the rows show plots for \textit{positive : negative} = \{10:1, 1:1, 1:10\} respectively. 
        From left to right, the columns represent epoch 1, 10, 100, 1000 and 10000. 
        The positive samples are colored with light blue, while the negative samples are colored with dark blue. 
        The background shading depicting the decision surface of an MLP. 
        The red lines are decision boundaries of Logistic Regression, and the dashed black lines are max-margin separators.
    }
    \label{fig:1}
\end{figure}

\begin{figure}[H]
    \centering
    \includegraphics[width= 0.85\textwidth]{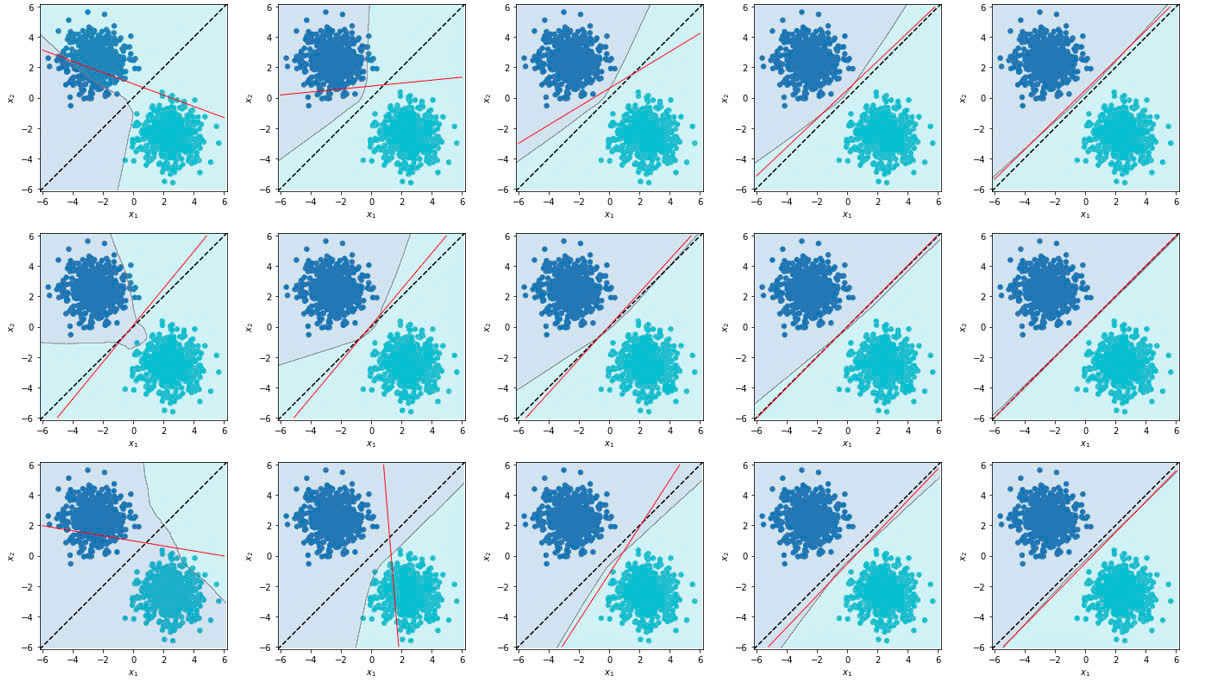} 
    \caption{
        Results on linearly-separable dataset. The model is LR and MLP with L2 regularization,
        without importance weighting and with loss function weighted by $1/r$. 
        We also train models with weighted loss function and L2 regularization. 
        We use the same CNN architecture as defined above and hyperparameters also hold the same. 
        For each experiment, we train 3 models with a different random seed to compute mean and standard deviation.
    }
    \label{fig:2}
\end{figure}

\begin{figure}[H]
    \centering
    \includegraphics[width= 0.85\textwidth]{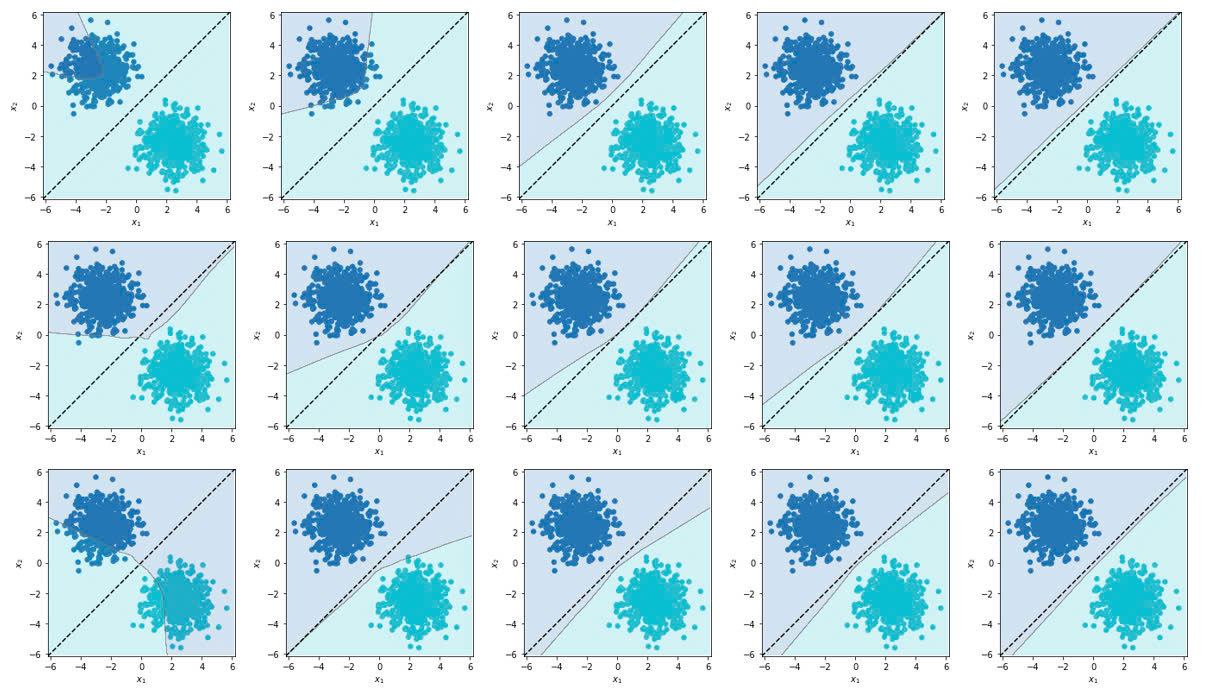} 
    \caption{
    The model is MLP with dropout rate = 0.5
    }
    \label{fig:3}
\end{figure}

\begin{figure}[H]
    \centering
    \includegraphics[width= 0.9\textwidth]{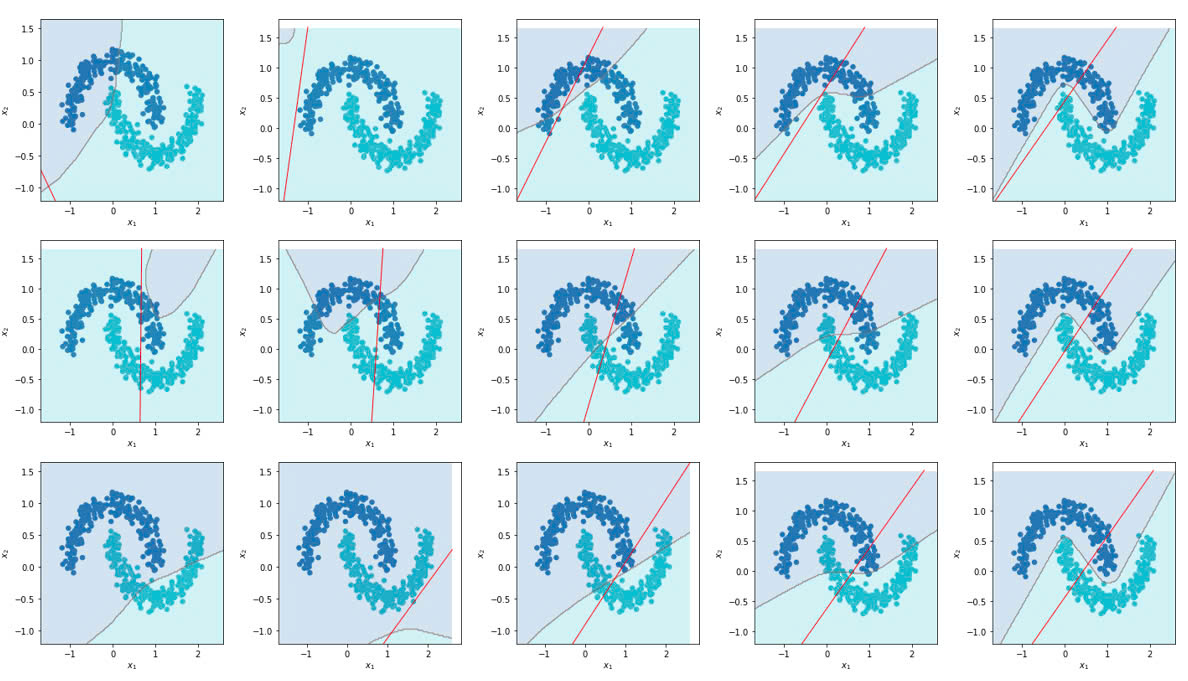} 
    \caption{
Results on not linearly-separable moons dataset. The models are LR and MLP without L2
regularization.
    }
    \label{fig:4}
\end{figure}

For the linear-separable dataset, Figure~\ref{fig:1}, \ref{fig:2}, \ref{fig:3} show that no matter what weights applied to the loss function, the decision boundary will eventually converge to the max-margin separator for both Logistic Regression and MLP. For the moon dataset, Figure~\ref{fig:4} shows the experiment results. Although the decision boundary of MLP converges and separates the two classes completely with different importance weights, we observe that the decision boundary of the Logistics Regression model depends on the importance weights in epoch 10000. With positive:negative = 10:1, the decision boundary tends to place all positive samples in one side by misclassifying more negative samples, and vice versa. With the equal importance weights, the decision boundary makes errors on the two classes equally. This is because the LR model cannot separate the moon data perfectly. When the model inevitably makes misclassifications, it shows partiality for the side with higher importance in loss function to minimize the weighted risk.

\begin{figure}[H]
    \centering
    \includegraphics[width=\textwidth]{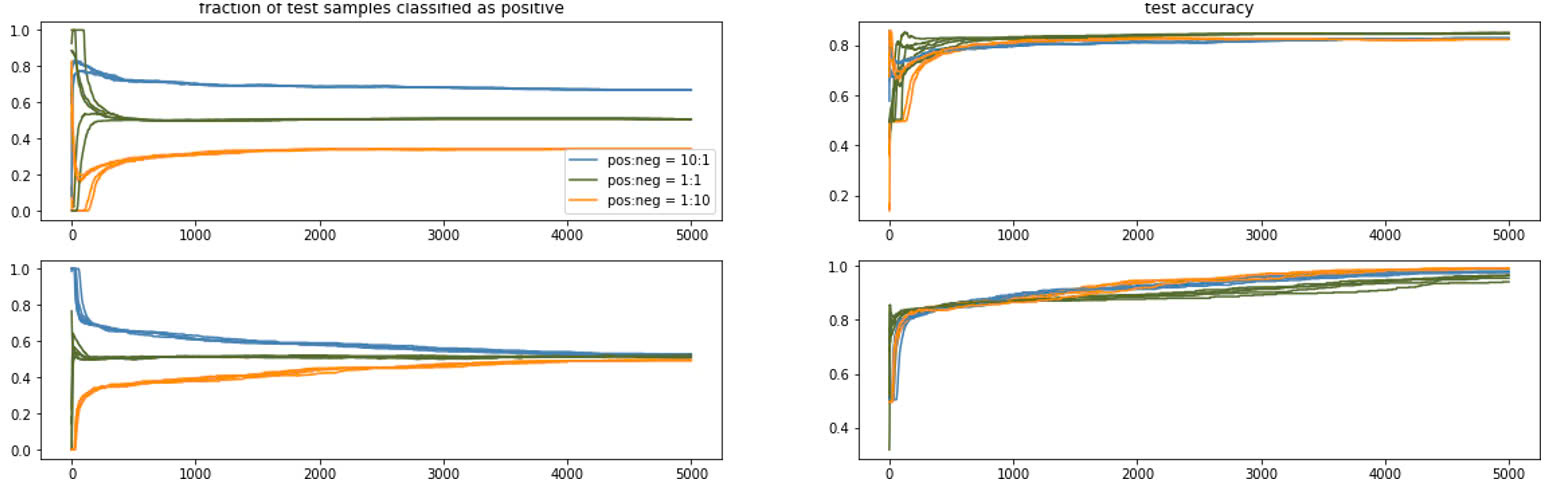} 
    \caption{
Results on not linearly-separable moons dataset. The top shows faction of positive prediction
and test accuracy against number of training epochs corresponding to LR, and the bottom shows
fraction and accuracy plots corresponding to MLP
    }
    \label{fig:5}
\end{figure}

We question whether the difference between the visualizations associated with epoch 1 is caused by down-weighting or up-weighting at the initialized state, and whether the convergence depends on the initialized state. To validate this, we repeat the experiments on moon data for 5 trails and record the tracks of the fraction of positive predictions and the test accuracy during the training process. In Figure~\ref{fig:5}, we observe that the initialized states are highly random, but the lines associated with the same weights gather rapidly. The final states at the end of epoch 10000 are related to importance weights for Logistic Regression while independent on importance weights for MLP.

\begin{figure}[H]
    \centering
    \includegraphics[width=\textwidth]{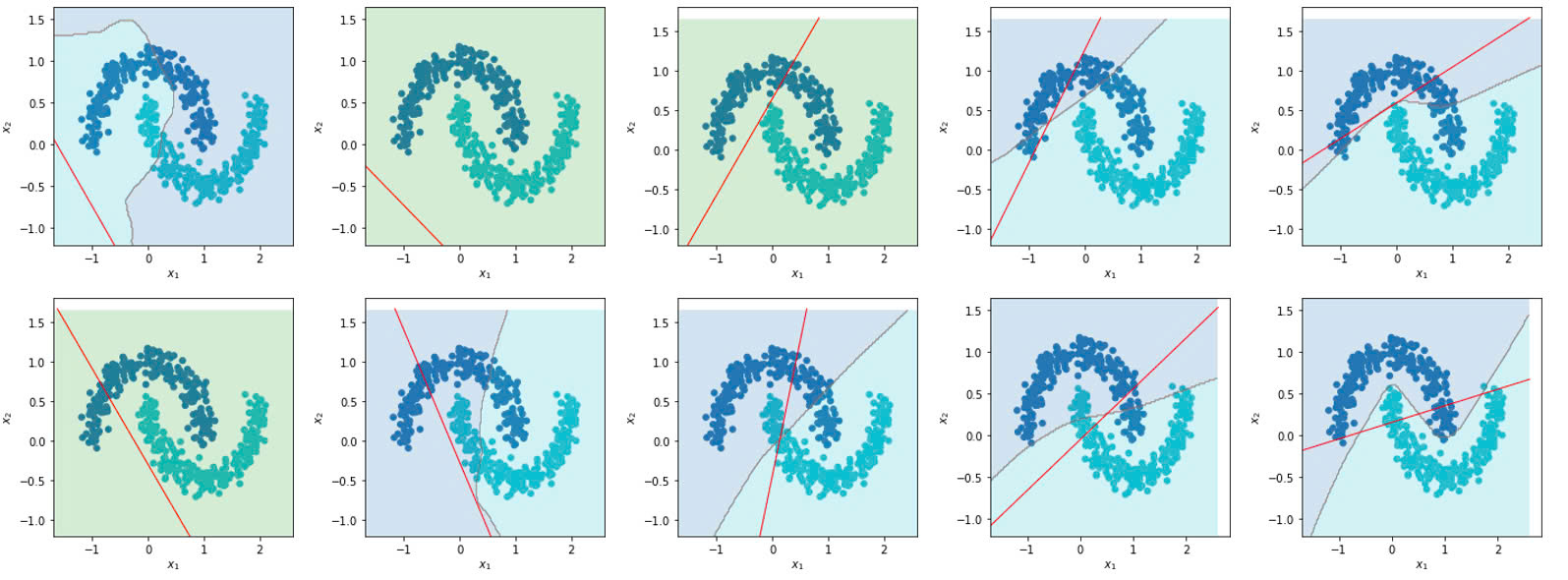} 
    \caption{
        Results on imbalanced moon dataset. In training set, \textit{positive : negative} = $10:1$. 
        The top shows the change of decision boundaries with unweighted loss function. 
        The bottom shows the change of decision boundaries with loss function up-weighting the negative samples by 10. 
        The green background shadows mean that the MLP classifies the whole surface as the same label.
    }
    \label{fig:6}
\end{figure}

\begin{figure}[H]
    \centering
    \includegraphics[width=\textwidth]{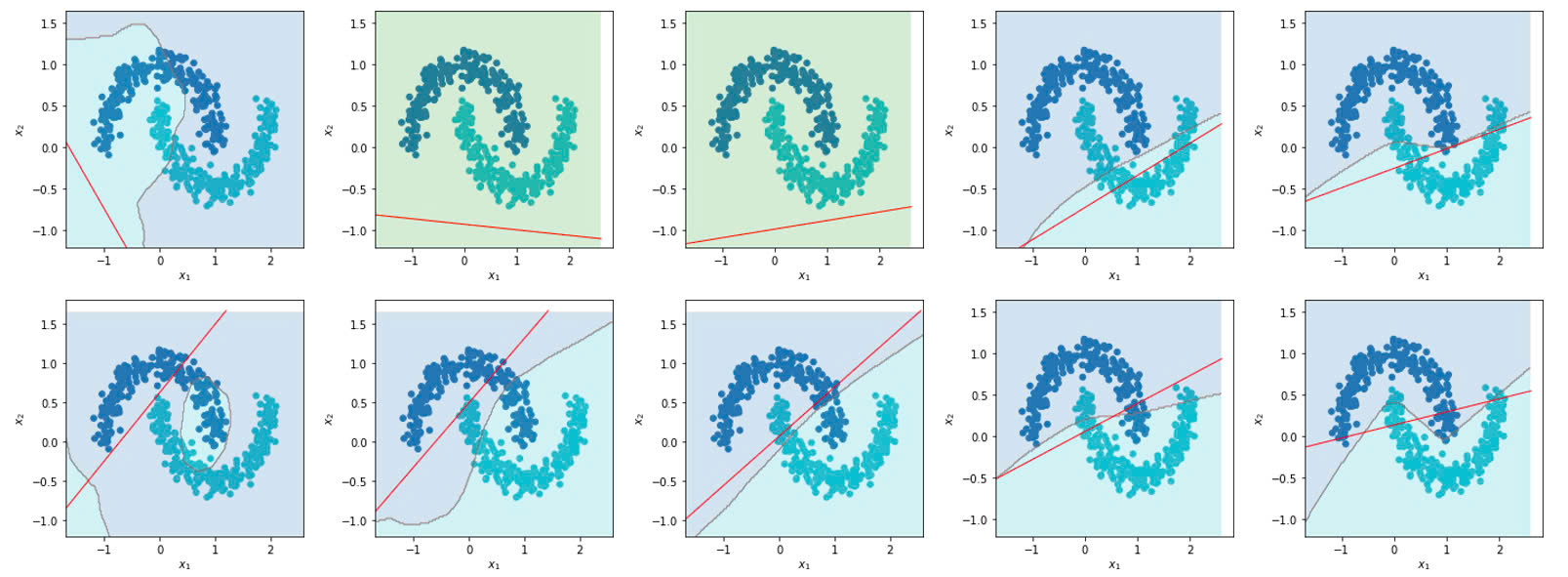} 
    \caption{
        Results on imbalanced moon dataset. In training set, \textit{positive : negative} = $1:10$. 
        The top shows the change of decision boundaries with unweighted loss function. 
        The bottom shows the change of decision boundaries with loss function up-weighting the positive samples by 10. 
        The green background shadows mean that the MLP classifies the whole surface as the same label.
    }
    \label{fig:7}
\end{figure}

Figure~\ref{fig:6} and \ref{fig:7} show the results for imbalanced moon data. We find that weighting the loss function according to the ratio of positive to negative labels in the training set can improve both the Logistic

Regression and the MLP. With weighted loss, the models can make predictions as accurately as they
are on the balanced training set, while with unweighted loss function, the decision boundaries show
partiality to the side with more training examples.

\subsection{CIFAR-10 Binary Classification}

\begin{figure}[H]
    \centering

    \begin{subfigure}[b]{0.3\textwidth}
        \includegraphics[width=\textwidth]{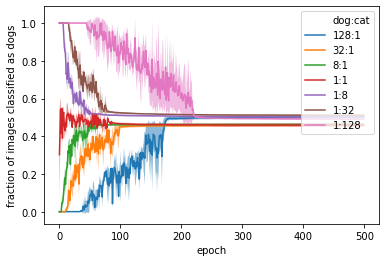}
        \caption{CIFAR-10 cat and dog test images.}
            \label{fig:8a}
    \end{subfigure}
    \hfill
    \begin{subfigure}[b]{0.3\textwidth}
        \includegraphics[width=\textwidth]{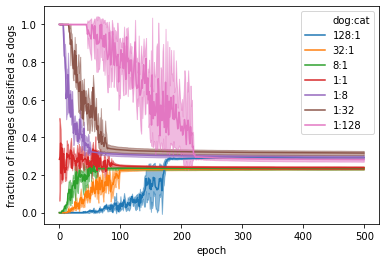} 
        \caption{CIFAR-10 other 8 classes test images.}
                 \label{fig:8b}
    \end{subfigure}
    \hfill
    \begin{subfigure}[b]{0.3\textwidth}
        \includegraphics[width=\textwidth]{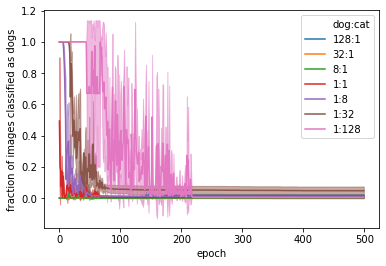} 
        \caption{Random images.}
                 \label{fig:8c}
    \end{subfigure}

    \vspace{1em} 

    \begin{subfigure}[b]{0.3\textwidth}
        \includegraphics[width=\textwidth]{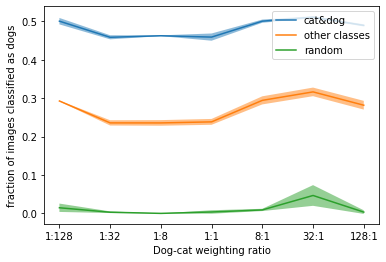} 
        \caption{Classification ratios at epoch 500.}
                \label{fig:8d}
    \end{subfigure}
    \hfill
    \begin{subfigure}[b]{0.3\textwidth}
        \includegraphics[width=\textwidth]{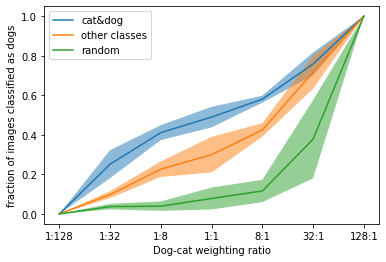} 
        \caption{Classification ratios with L2 regularization.}
                \label{fig:8e}
    \end{subfigure}
    \hfill
    \begin{subfigure}[b]{0.3\textwidth}
        \includegraphics[width=\textwidth]{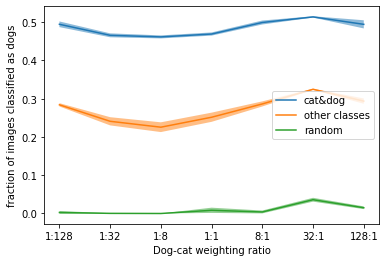} 
        \caption{Classification ratios with dropout.}
                \label{fig:8f}
    \end{subfigure}

    \caption{
        (a-c) Relationship between early stopping and importance weighting. We plot the fraction of images classified as dogs (y-axis) in the cat and dog test set (a), other eight classes (b), and random images (c) vs training epochs (x-axis). 
        (d-f) Fraction of examples classified as dogs (y-axis) vs importance weights (x-axis) after 500 epochs of training. We also show results from models trained with L2 regularization (e) and dropout (f). In all plots error bands show standard deviation across three random initializations, and lines represent means.
    }
    \label{fig:8}
\end{figure}

Training epochs on different important weights are plotted. In Figure~\ref{fig:8a}, we observe that even though during early epochs, the fractions of images classified as dogs correspond to the importance weights, the discrepancy keeps diminishing. When the epoch reaches to around 230, the fractions of dogs on all important weighted models converge to 0.5, which is equal to the fraction of dogs in the cat and dog test set. Similar patterns are discovered among the test images of other eight classes in Figure~\ref{fig:8b} and random noise images in Figure~\ref{fig:8c}. When epoch reaches to around 230, the fractions of images predicted as dogs for all weighted models converge to around 0.3 among other 8 classes and converge to near 0 among random images. The model with different weighting ratios agrees on the out-of-sample images. Besides, it seems that all models have near-perfect agreement on random noise images which are nearly-always classified as cats. These three figures show that as training progresses, the effects due to importance weighting vanish. The weighting impacts the CIFAR-10 binary classification early in training, however, after many epochs of training, there is no

\subsection{CIFAR-10 Binary Imbalanced}
\begin{figure}[ht]
    \centering
    \subfloat[Without importance weighting\label{fig:9a}]{
        \includegraphics[width=0.3\textwidth]{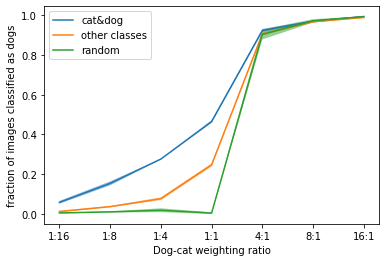}
    }
    \hfill
    \subfloat[With importance weighting\label{fig:9b}]{
        \includegraphics[width=0.3\textwidth]{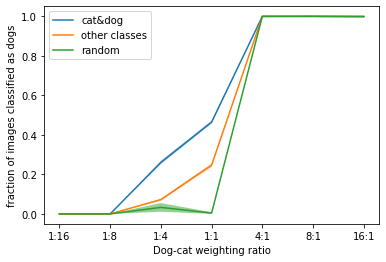}
    }
    \hfill
    \subfloat[With weighting and add L2\label{fig:9c}]{
        \includegraphics[width=0.3\textwidth]{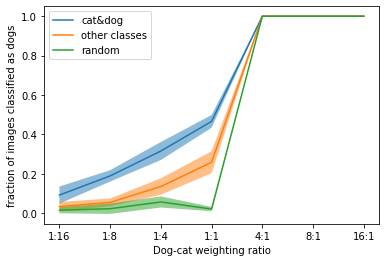}
    }
    \caption{Imbalanced binary classification with and without importance weighting}
    \label{fig:9}
\end{figure}

Importance weighting is commonly used to correct for class imbalance. Figure~\ref{fig:9} shows the results of training imbalanced data without weighting, with weighting and with weighting along with L2 regularization. Figure~\ref{fig:9a} indicates that the imbalanced training data makes the models more incline to predict images as the class with more training data. However, as demonstrated in Figure~\ref{fig:9b}, importance weighting does not help with class imbalance in this case. Moreover, adding L2 regularization does not make any difference neither.

\subsection{Covariate Shift}

\begin{figure}[ht]
    \centering
    \includegraphics[width=0.45\textwidth]{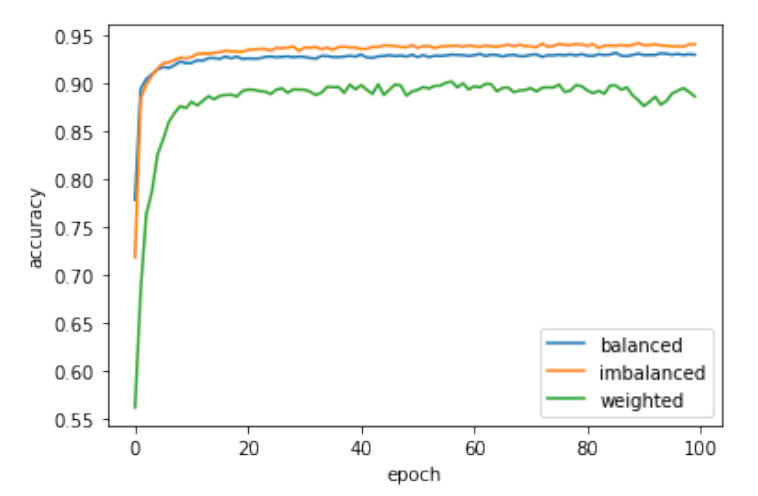}
    \hfill
    \includegraphics[width=0.45\textwidth]{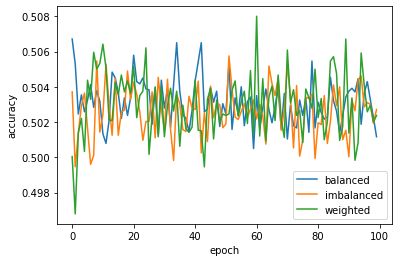}
    \caption{Results on dataset with covariate shift with train on the left and test on the right.}
    \label{fig:10}
\end{figure}

By combining two original classes into a new class twice, we were able to explicitly form the desired covariate shift and calculate the optimal importance weights, but at the same time, under each new class, we created a heterogeneous dataset, making it challenging for our models to extract common patterns and produce well-generalized predictions. Therefore, as illustrated by the right plot in Figure~\ref{fig:10}, the validation accuracies for three methods were around 0.503. Due to the heterogeneity under each new class, our models actually optimized by relying on features of only one particular group, e.g. either dog or cat, of each class to make classifications. Therefore, whether or not we chose to downsample one particular group did not bring a huge impact on the performances of the models. The importance weights, however, forced the model to pay similar attentions to both groups in a class, leading to the decreased training accuracy as indicated by the gap between the green and the blue/orange lines (Figure~\ref{fig:10}). Had there been less heterogeneity within the new classes and more between, we could expect the weighted model successfully capturing features of both groups of each new class to outperform the unweighted model and gradually approach the optimal accuracy achieved when there is no covariate shift between training and testing datasets.

\section{Conclusions and Discussion}

In this project, we discuss the effect of importance weighting in different scenarios thoroughly. With the synthetic data, we show the importance of weighting affects the decision boundary in early state, but along with the increasing number of training epochs, the impact vanishes and depends on datasets and models at the final states. When data is linearly separable, simple machine learning models can easily make predictions with 100\% accuracy, so the effect of importance weighting in loss function diminishes and the decision boundaries converge to the max-margin separators (Figure~\ref{fig:1}). When data is not linearly separable, the impact is different for misspecified model and correctly specified model (Figure~\ref{fig:4}). A linear model (LR) trained on the inseparable data classifies more test examples as the label assigned higher weights at the final states. A non-linear model (MLP), as is correctly specified, can separate the two classes perfectly no matter how the loss function is weighted.

In the case of training deep neural networks for binary image classification on CIFAR-10, we discover that, similar to the paper~\cite{ref1}, the importance weighting effects the predictions early in training. However, the effect dissipates after training for enough epochs and all the weighted models converges to similar prediction ratios. We suspect that since the deep learning models are over-parameterized and very expressive, they can usually separate the classes after training for appropriate number of epochs regardless of the weightings. Therefore, deep nets approach similar solutions across different weighted models on either test set images, out-of-domain images, or random vectors in our experiments, after several hundreds of training epochs.

To test the impact of importance weights when combining with regularization techniques, we use L2 regularization and dropout. With the linearly separable training set, there is little difference in epoch 10000 over the decision boundaries when applying L2 regularization or dropout or not either Figure~\ref{fig:1}, \ref{fig:2}, \ref{fig:3}. This is reasonable as our linearly-separable dataset is so simple that the regularization techniques are not effective on the results. However, we notice there are obvious impacts of importance weights when combining L2 regularization, but not dropout, to deep nets on the CIFAR-10 dataset. This is consistent with the conclusion as described in~\cite{ref1} that L2 regularization restores certain effects of importance weighting. Intuitively, it makes sense that importance weighting works better with more regularization.

 To test the impact of weight correction on imbalanced data, we sub-sample one class in the training set while keeping the testing set distributed evenly on the two classes. With synthetic imbalanced moon data (Figure~\ref{fig:6}, \ref{fig:7}), both the unweighted linear model (LR) and unweighted non-linear model (MLP) classify all test samples in the class with more training samples correctly but make mistakes on another class. But when using importance weights to correct the imbalance, the models improve a lot and classify the test set almost evenly in epoch 10000, and the non-linear model can almost perfectly separate the two classes. This indicates the effectiveness of importance weighting on imbalance correction when using traditional machine learning models. When classifying CIFAR-10 imbalanced data, deep neural networks, on the other hand, do not manifest any differences before and after using weighted loss function to combat the class imbalance, which is consistent with the appendix 6 of the paper~\cite{ref1}. However, surprisingly, contrary to the conclusion we made on CIFAR-10 binary classification, L2 regularization does not restore any effects of importance weighting in the case of class imbalance. More experiments are needed to make rigorous conclusions from this conflicting results. We suspect one possible reason might be that we have different total training samples for different sub-sampled training sets.

We don’t see much difference among the test accuracies in the three experiments we conducted for the covariate shift problem. It seems that whether we train the model with covariate shift exists between train and test sets or not, the self-defined CNNs can’t generalize well on the test set. The importance weighting might play a role to some extent early in training, but the increase is too subtle to make solid conclusions from it. The possible reason could be the train and test dataset we create is not suitable for examining the covariate shift problem. In our course, the sample data point is 1-D dimension, but the image data is 2-D dimension. If we have more time, we may leverage other more reasonable dataset to test the effect of importance weight in the covariate shift problem. We may also apply more sophisticated models with importance weighting to enhance the model performance.

\end{document}